\documentclass[letterpaper]{article} 
\usepackage{aaai2026}  
\usepackage{times}  
\usepackage{helvet}  
\usepackage{courier}  
\usepackage[hyphens]{url}  
\usepackage{graphicx} 
\usepackage{natbib}  
\usepackage{caption} 
\usepackage{algorithm}
\usepackage{algorithmic}

\usepackage{newfloat}
\usepackage{listings}
\DeclareCaptionStyle{ruled}{labelfont=normalfont,labelsep=colon,strut=off} 
\floatstyle{ruled}
\newfloat{listing}{tb}{lst}{}
\floatname{listing}{Listing}
\usepackage{comment}
\usepackage{subcaption}
\usepackage{booktabs}
\usepackage{multirow}
\usepackage{placeins}
\usepackage{etoolbox}

\title{Predicting Groundwater Arsenic Concentrations Using Graph Neural Networks}

\author{
    William Xing, 
    Stephanie Yang, 
    Aarush Bandemegal, 
    Anushree Misra, 
    Ananya Kalapatapu, 
    Brennan Lagasse,
    Kevin Zhu
}
\affiliations{
    Algoverse AI Research\\
    \{williamlelexing, stephanieyang1107, aninja2026, shreemisra23, ananya.kalapatapu\}@gmail.com, \{brennan, kevin\}@algoverseairesearch.org
}

\begin{document}

\maketitle

\begin{abstract}
Arsenic contamination in groundwater presents a longstanding public health crisis in the United States, especially for households depending on private wells. Accurate and spatially informed prediction of arsenic concentration is vital to identify high-risk areas and focus mitigation efforts. However, there is a lack of generalizable models for representing continuous variation in arsenic concentrations across regions. In this work, we pose arsenic prediction as a regression task and construct a spatially integrated dataset to aggregate over 74,000 arsenic samples from the Water Quality Portal (WQP), Mineral Resources Data System (MRDS), and Gridded National Soil Survey Geographic Database (gNATSGO). Specifically, we use a variety of techniques including k-Nearest Neighbors (k-NN) and Geographic Information Systems (GIS) to join arsenic measurement points from across the United States by location. Building on this dataset, we evaluate a diverse suite of machine learning models, including tree-based ensemble approaches, multilayer perceptrons, and spatially aware graph neural networks (GNN). Our findings show that while gradient-boosted trees are still considered state-of-the-art in the field of tabular data, GNNs are able to further account for spatial dependence to match or outperform the results of gradient-boosted trees. These results demonstrate that graph-based and spatially informed learning can enhance environmental prediction and provide a foundation for improved groundwater risk mapping and monitoring.
\end{abstract}

\begin{links}
    \link{Code}{https://github.com/BrennanLagasse/Water-Contamination-Prediction}
\end{links}

\section{Introduction}

Arsenic contamination in groundwater is currently a major public health issue in the United States. 2.1 million Americans are estimated to drink water from wells that have dangerously high arsenic levels \cite{Ayotte2017EstimatingTH}. Both the World Health Organization (WHO) and the Environmental Protection Agency (EPA) have set the safe limit for arsenic at 10 \textmu g/L, but numerous regions in America, particularly in states such as Nevada and Arizona, exceed this amount \cite{Lombard2021MachineLM}. Chronic exposure to arsenic can lead to cardiovascular disease, diabetes, and cancer. Although direct chemical testing is the most reliable way to assess arsenic levels, this option is often infeasible due to financial or logistical constraints, especially for people who live in rural areas, where arsenic contamination is more prevalent and reliance on private wells is more common.

Recent studies have applied machine learning to address the prediction of arsenic contamination in groundwater, but several challenges remain to achieve robust generalization and real-world deployment. Very few prior studies approach arsenic prediction as a regression task, despite the fact that regression would allow for the creation of much more detailed prediction maps of areas with poor water quality \cite{Hu2022TheUO}. For example, two samples containing 9.9 \textmu g/L and 10 \textmu g/L of arsenic are nearly identical and both indicate unsafe conditions, yet a classification model would label one as "safe" and the other as "unsafe" while not providing further information.

While previous studies have reported very low loss metrics such as mean absolute error (MAE) or mean square error (MSE), many have focused on smaller or region-specific datasets. These approaches provide valuable localized insights but are often limited in their ability to generalize across diverse geologic settings. Moreover, there are studies that conduct experiments on a national or global scale, but their lack of data density remains a critical barrier, as arsenic values can vary greatly across short distances due to factors such as depth differences and geochemistry. Lastly, very few previous papers have attempted to predict arsenic concentrations in water using machine learning techniques in the United States, instead opting to focus more on various South American and Asian countries.

In this study, we aim to address this critical gap in the literature and introduce two novel advancements in the study of the prediction of arsenic contamination. First, we provide a unified dataset for arsenic prediction which unifies location and geological datasets for a total of 74,706 data points. This dataset includes a diverse set of features that influence arsenic contamination and provides sufficient sample density to represent the corresponding geographic area. Our analysis focuses on the contiguous United States, which, compared to previous studies of localized or homogeneous areas, has not yet been extensively examined in detail by itself as an independent modeling region. Second, we train and evaluate more modern machine learning models such as graph neural network (GNN) variants for arsenic prediction alongside classical approaches such as random forests and gradient-boosted trees. Although GNNs have not yet been applied to water contamination prediction, their ability to model spatial autocorrelation makes them particularly well-suited for this task, an important challenge highlighted in \cite{Hu2022TheUO}. This ability allows for predictions to be made about a certain point using information from its surrounding points, which is necessary in the case of arsenic contamination. We are also concerned that the tree-based models widely used in prior studies may not fully capture the underlying relationships between contamination levels and environmental features, particularly due to their false assumption that each data point is independent of the others and their reliance on axis-aligned splits. Although tree-based ensemble models remain highly competitive, we found that advanced variations of graph neural networks (GNN) are able to outperform them, highlighting their growing potential for environmental science applications. Using a dense, geographically diverse dataset that spans the continental U.S, we demonstrate improved model performance and enhanced applicability for large-scale arsenic risk prediction.

\section{Related Works}
Recent research has predominantly approached arsenic prediction as a classification problem, with one survey of 27 drinking water studies reporting that 70\% of the reviewed papers adopted this approach \cite{Hu2022TheUO}. Furthermore, the majority of these works used tree-based ensemble models, presumably because of their robustness and well-established performance on tabular data. For example, \cite{Podgorski2020GlobalTO} attempts to predict global arsenic concentrations using around 55,000 data points, although their data is mostly concentrated in Asia, North America, and Europe. They aggregate all data points within 1-$km^2$, only keep samples from above 100 m depth, and use a random forest to make predictions. \cite{Wu2021DistributionOG} models arsenic concentrations in groundwater across Uruguay, introducing a hybrid approach by combining data points at different depths or geological compositions. They apply spatial aggregation by aggregating arsenic measurements within the same pixel using the geometric mean of values. Various papers attempt to model arsenic contamination in India, a country known for its water pollution. The studies focused on India utilize boosted regression trees or random forest models for prediction and data processing techniques such as kriging or spatial analysis. \cite{Chattopadhyay2023TheML, Tan2020MachineLM, Podgorski2020GroundwaterAD, Wu2023ArtificialIM}. 

Not many studies have applied regression models to predict water contamination, much less specifically for arsenic concentration. Tree based methods are still very popular for regression. \cite{Subudhi2025IntegratingBL} predicts the groundwater quality index in Odisha and also uses feature importance to find the most effectual contaminants to this value. Their dataset consists only of samples from 2019-2022 and includes features such as pH, electrical conductivity, and other substances found in groundwater. As for the specific machine learning algorithms, they use gradient-boosted trees including XGBoost, LightGBM, and CatBoost along with random forests. \cite{Zhao2024PredictingAC} employs multiple linear regression (MLIR), multivariate logistic regression (MLOR), and random forest models to predict arsenic concentrations in groundwater in the Hetao Basin (China) and several regions of Bangladesh. Among these, the random forest model outperforms MLIR and MLOR.

Although GNNs have not been widely applied to arsenic prediction, they have been frequently used in other environmental science domains, where their capacity to capture spatial correlations makes them particularly effective. For example, \cite{Zha2024GNN} uses GNN variants to predict the amount of heavy metals in the Pearl River Basin, China. Their GNN graph was created by first connecting nodes based on geographic distance and further refined by creating a reachability matrix to connect remaining nodes and avoid having a non-connected graph. They also found that GNNs are able to handle high-dimensional data and reveal complex spatial patterns of soil heavy metal distribution. \cite{Truong2023GraphNN} combines GNNs with physics-based modeling to estimate pressure and flow in water distribution networks. Their nodes represent the endpoints of the pipes (junctions, tanks, reservoirs, etc.) and their edges correspond with real-world pipe connections, modeling the relations between these endpoints. \cite{Sun2022AGN} similarly introduces a physics-guided GNN approach for basin-scale river network learning and streamflow forecasting.

\section{Methodology}

\subsection{Data Sources}
We created our data set by merging other datasets from a variety of sources. The Water Quality Portal (WQP) \cite{environmental_protection_agency_water_2013} is a reputable water contamination data source, integrating data from the United States Geological Survey (USGS), the Environmental Protection Agency (EPA), and other state, federal, tribal, and local agencies. In addition to nationwide water contamination measurements, the dataset includes predictive variables such as latitude, longitude, and well depth. To incorporate soil properties, we used the Gridded National Soil Survey Geographic Database (gNATSGO) \cite{soil_survey_staff_gridded_2025}, which provides high-resolution spatial data on soil horizons, components, and other attributes relevant to contaminant mobility and retention. Some of the predictive data provided by this dataset include the percentage of elements like clay and organic matter in the soil, the pH of the soil, and the slope gradient and elevation of the surrounding environment. Geological and mineralogical data were added from the Mineral Resources Data System (MRDS) \cite{mason_mineral_1996}, which catalogs mineral deposit locations and a variety of associated parameters. Some examples of these parameters are the minerals present, geomorphic location, and the physiographic division.

\begin{figure*}[tb]
    \centering
    \begin{subfigure}[b]{0.32\linewidth}
        \includegraphics[width=\linewidth]{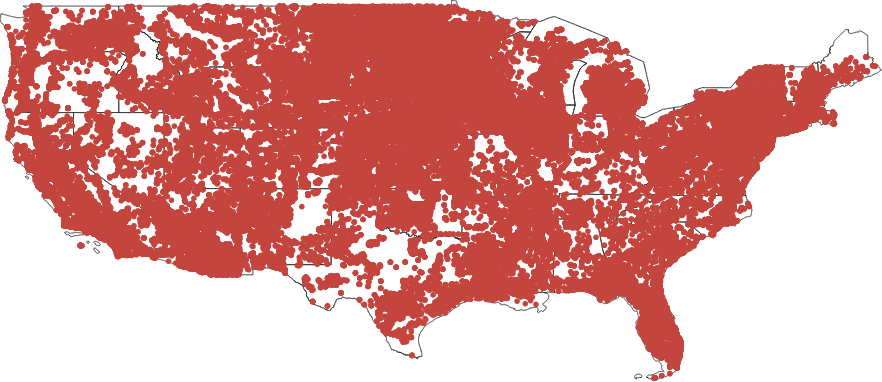}
        \caption{WQP Stations}
        \label{fig:wqp-map}
    \end{subfigure}
    \hfill
    \begin{subfigure}[b]{0.32\linewidth}
        \includegraphics[width=\linewidth]{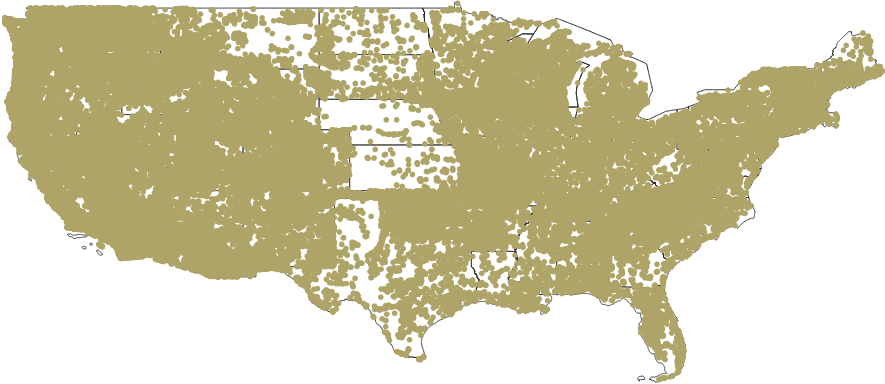}
        \caption{MRDS Sites}
        \label{fig:mrds-map}
    \end{subfigure}
    \hfill
    \begin{subfigure}[b]{0.32\linewidth}
        \includegraphics[width=\linewidth]{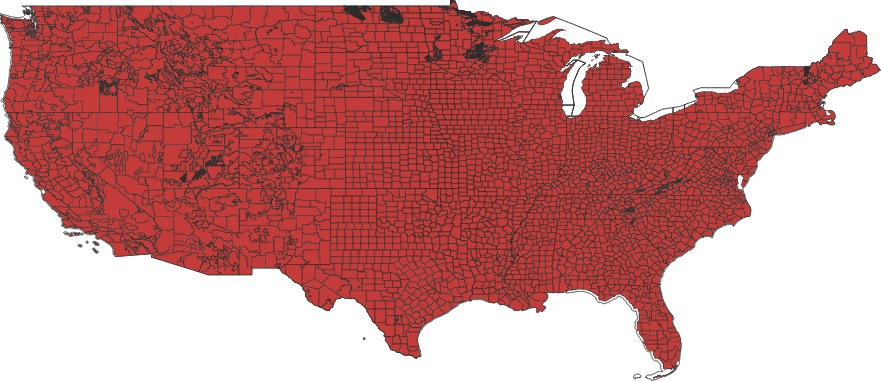}
        \caption{gNATSGO Soil Polygons (Represent mapped soil units)}
        \label{fig:gnatsgo-map}
    \end{subfigure}
    \caption{Spatial coverage of datasets used for arsenic prediction across the mainland United States.}
    \label{fig:arsenic-three}
\end{figure*}

Together, these datasets offer a complementary view of the environmental conditions that impact water contaminant levels. We provide the geographic distribution of each dataset in Figure~\ref{fig:arsenic-three}.

\begin{figure}[tb]
    \centering
    \includegraphics[width=1\linewidth]{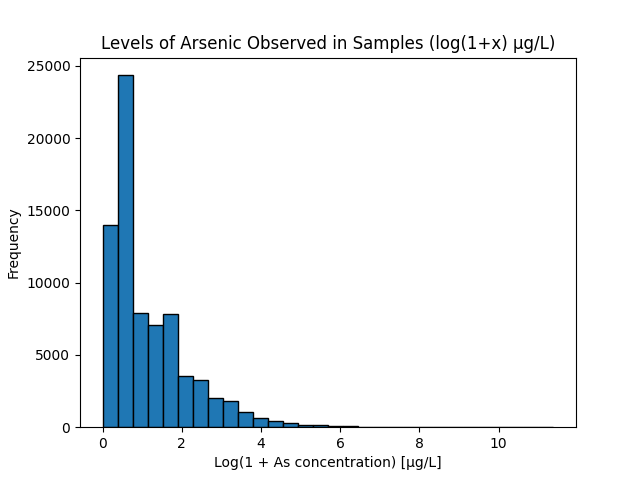}
    \caption{Distribution of arsenic levels in log scale}
    \label{fig:arsenic_level}
\end{figure}

Our dataset has 74,706 samples, each corresponding with an arsenic measurement, and 115 features describing the location of those samples. Approximately 87.69\% of samples have a safe level of arsenic, and about 12.31\% of samples have an unsafe level of arsenic. Figure~\ref{fig:arsenic_level} illustrates a detailed distribution of arsenic concentrations across the country.

\subsection{Feature Engineering}
\begin{figure}[tb]
    \centering
    \includegraphics[width=1\linewidth]{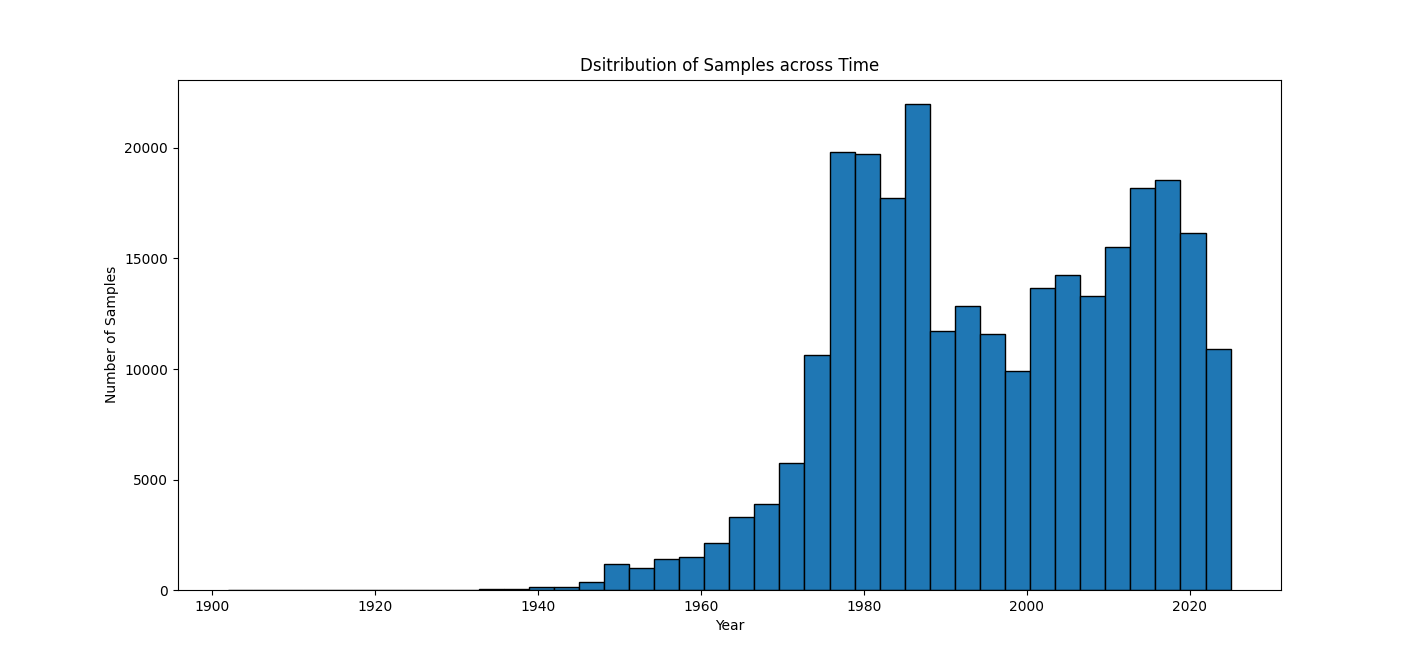}
    \caption{Distribution of samples across time}
    \label{fig:arsenic_across_time}
\end{figure}

We only kept the most recently collected arsenic samples from each WQP station to avoid data leakage and temporal bias. Figure~\ref{fig:arsenic_across_time} represents the distribution of arsenic samples across time after filtering out older data samples.
When arsenic values were missing, detection limits were typically reported, and when arsenic values were available, detection limits were usually absent. Because of this strong correlation between missing arsenic values and the presence of detection limits, we imputed the arsenic value with half the detection limit. This covered over 97\% of missing values. We also filled in missing values in the "Well Depth" column by grouping together rows by their corresponding county code then imputing the median for each county. These two techniques boost the number of training values available, allowing for more accurate model performance.

\begin{table*}[tb]
\caption{Comparison of model performance across log-scale and real-scale metrics.}
\centering
\begin{tabular}{llcccccc}
\toprule
\textbf{Model Type} & \textbf{Model Name} &
\multicolumn{2}{c}{\textbf{MAE (\textmu g/L)}} &
\multicolumn{2}{c}{\textbf{MSE (\textmu g/L$^2$)}} &
\multicolumn{2}{c}{\textbf{$R^2$}} \\
\cmidrule(lr){3-4} \cmidrule(lr){5-6} \cmidrule(lr){7-8}
& & \textbf{Log} & \textbf{Real} & \textbf{Log} & \textbf{Real} & \textbf{Log} & \textbf{Real} \\
\midrule
\multirow{1}{*}{Simple Tree Models}
 & Random Forest & 0.538 & 3.966 & 0.605 & 70.032 & 0.439 & 0.298 \\
\midrule
\multirow{1}{*}{Simple Neural Networks}
 & Multilayer Perceptron & 0.563 & 3.623 & 0.739 & 223.610 & 0.292 & -1.187 \\
\midrule
\multirow{2}{*}{Gradient-Boosted Trees}
 & XGBoost & 0.527 & 3.399 & 0.675 & 83.958 & 0.374 & 0.158 \\
 & LightGBM & 0.538 & 3.971 & 0.627 & 72.894 & 0.418 & 0.269 \\
 & CatBoost & 0.533 & 3.433 & 0.680 & 87.133 & 0.369 & 0.126 \\
\midrule
\multirow{4}{*}{GNN Variants}
 & GCN & 0.528 & 3.485 & 0.701 & 86.381 & 0.382 & 0.176 \\
 & GAT & 0.522 & 3.319 & 0.682 & 75.039 & 0.370 & 0.215 \\
 & GraphSAGE & 0.529 & 3.256 & 0.696 & 72.981 & 0.332 & 0.191 \\
 & GIN & 0.546 & 3.282 & 0.759 & 76.801 & 0.326 & 0.201 \\
\bottomrule
\end{tabular}
\label{tab:comparison}
\end{table*}

For the MRDS dataset, we grouped together samples across time from the same MRDS station. One important technique we used here was word embeddings. MRDS contained many non-numeric columns, such as main commodity found and physiographic sector. We mapped these words to vectors of length 2, 4, or 8 using PyTorch embeddings. Lastly, we joined this physiographic information back to the original WQP dataset by averaging the values for rows within 5 km of a WQP station. In the case where no nearby stations within the set radius are found, the all values for that station are kept as 0. All coordinates were standardized to WGS84 (EPSG:4326), and distances were computed using the haversine metric. The low distance of 5 km ensures that the mineral information is relevant to the arsenic sampling location and the default value of 0 instead of assigning no values ensures that rows that are still useful for training.

gNATSGO is a bit different from the other datasets in that it divides the United States into different sectors instead of taking samples from specific locations. As a result, we used GIS to determine which gNATSGO sector contained each WQP station. Each sector has its own id, called a MUKEY, and each sector contains multiple components that also have an id, called a COKEY. Each COKEY corresponds to specific soil data such as pH levels and the concentration of different particles (sand, silt, etc.). Because gNATSGO divides the United States into polygonal map units rather than point-based sampling locations, we could not directly join this dataset to our WQP records using a k-nearest neighbors approach. Instead, we visualized both the WQP stations and gNATSGO polygons in a geographic information system (GIS) and used spatial overlay operations to assign each WQP station the corresponding MUKEY identifier. Because each station had one MUKEY but each MUKEY was mapped to multiple COKEYs, the rows were finalized by averaging the numeric values. For rows without a MUKEY, we used k-nearest neighbors (k-NN) to impute the MUKEY and all the data of those rows to its nearest neighbor. One hot encoding was used for columns that did not contain too many different values and embeddings were used for columns that did contain a range of different values. Similarly to handling the WQP data set, we imputed the median for numeric columns with few missing values.

\subsection{Predictive Models}
We first evaluated simple and widely used models to establish reference performance, then expanded to include advanced architectures. As a benchmark, we used random forests and multilayer perceptrons, two widely adopted ensemble methods known for performing strongly on tabular data. We further assessed many types of graph neural networks (GNN), including graph convolutional networks (GCN) \cite{Kipf2016SemiSupervisedCW}, graph attention networks (GAT) \cite{Velickovic2017GraphAN}, GraphSAGE \cite{Hamilton2017InductiveRL}, and graph isomorphism networks (GIN) \cite{Xu2018HowPA}. For additional comparison, we compared a range of gradient-boosted trees, including XGBoost \cite{Chen2016XGBoostAS}, LightGBM \cite{Ke2017LightGBMAH}, and CatBoost \cite{Ostroumova2017CatBoostUB}. Although GNNs have been rarely applied in past water contamination studies, their successful use in related environmental science fields and their proven ability to capture spatial dependencies make them a highly promising approach. 

At a high level, GNNs make inferences by collecting the feature vectors of each of its neighbors and aggregating this information with their own features. Neighbors are defined by the edges that are established during the creation of the graph. This kind of message passing allows for information to continuously flow further in the graph (i.e. after 2 passes a node's embedding will reflect information from its neighbors and its neighbors' neighbors). This allows for information to propagate through multiple passes and generalize to unseen regions, an important property for large-scale datasets. After this process is complete, all the embeddings are fed into a small neural network to produce a prediction. Different GNN variants mainly vary in how they aggregate neighbor information.

The way the graph is constructed plays a crucial role in determining GNN performance, as it controls how information propagates between nodes. In our setup, each node represents a sampling location defined by its latitude and longitude, and edges were constructed using the k-nearest neighbors approach based on spatial proximity. Because we only kept the most recent samples in the dataset, no two data points are from the same node. We experimented with multiple neighborhood sizes (e.g., 5, 10, 15, 20) and found that k=5 consistently yielded the best results across all models. The k-nearest neighbor graph captures local spatial dependencies in arsenic concentrations while preserving meaningful geographic structure. This design implicitly encodes spatial autocorrelation, a fundamental property of environmental processes, without introducing excessive noise from distant locations.

Each GNN variant was chosen for its unique strengths and modeling capabilities. GCNs update each node's representation by averaging and transforming the features of its neighboring nodes, effectively capturing local context. GAT introduces attention coefficients to weight each node differently, allowing this model to focus more on influential neighbors. GraphSAGE only samples a fixed amount of neighbors and uses a small neural network to aggregate them during training, allowing this model to better capture dependencies between neighbors. GIN also aggregates information from neighbors but does so using a much more expressive update rule, enabling it to capture subtle structural differences. Because environmental datasets often vary in spatial density, feature correlations, and noise structure, different GNN architectures will capture different aspects of the data. We therefore evaluate multiple variants to understand which modeling assumptions best align with the spatial and geochemical structure of arsenic contamination.

\begin{figure*}[tb]
    \centering
    \begin{subfigure}[t]{\columnwidth}
        \centering
        \includegraphics[width=\linewidth]{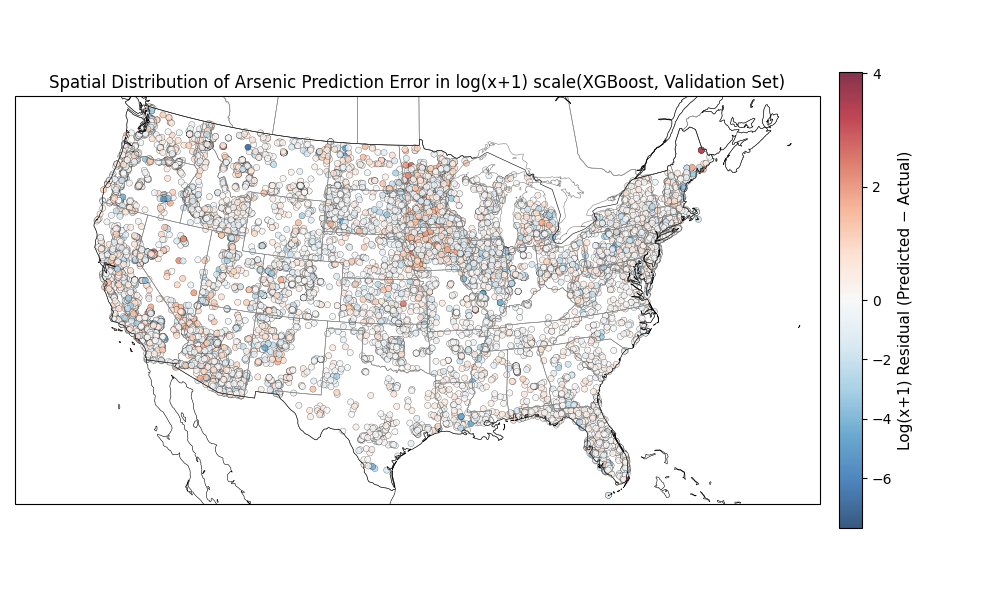}
        \caption{Spatial distribution of arsenic prediction error (XGBoost).}
        \label{fig:spatial_error_xgboost}
    \end{subfigure}
    \hfill
    \begin{subfigure}[t]{\columnwidth}
        \centering
        \includegraphics[width=\linewidth]{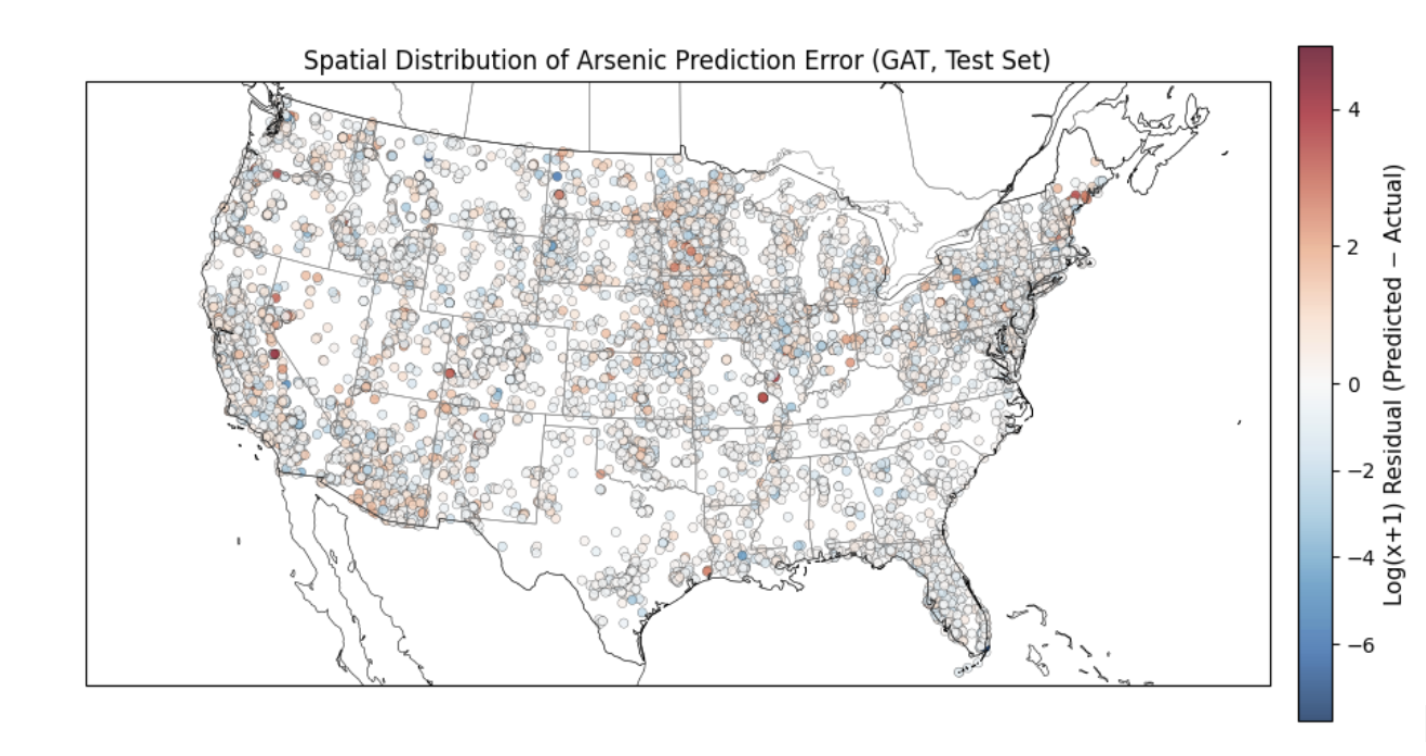}
        \caption{Spatial distribution of arsenic prediction error (GAT).}
        \label{fig:spatial_error_gat}
    \end{subfigure}
    \caption{Spatial distribution of error in prediction of log-levels of arsenic for the XGBoost and GAT models}
    \label{fig:gat_combined}
\end{figure*}

\section{Experiments}
We conducted experiments in both log(x + 1)-scale and real-value scale. The log(x + 1)-scale evaluation assesses the model's ability to predict the magnitude of arsenic contamination, while the real scale prediction generates results suitable for direct application. In many scenarios, being able to accurately predict the magnitude of arsenic contamination is extremely important due to how the safety limit (10 \textmu g/l) coincides with a power of 10 and the fact that arsenic concentrations are extremely skewed. In many environmental tasks, relative error often matters more than absolute error. For example, predicting 20 \textmu g/L instead of 10 \textmu g/L is more serious than predicting 110 µg/L instead of 100 µg/L. We applied a log(x + 1) transformation instead of a standard log(x) transform to handle samples with zero or very low arsenic concentrations. This approach avoids the undefined value at log(0) and prevents small concentrations from being mapped to disproportionately large negative values. By shifting the data by 1 µg/L before taking the logarithm, we preserve the relative differences among low values while compressing higher concentrations, resulting in a smoother and more stable target distribution for regression. When testing in real-value scale, we chose to remove all outliers above 100 $\mu$g/L as these outliers only accounted for around 1\% of the dataset and heavily inflated the loss. These few large values unnecessarily contribute disproportionately huge gradients to the loss, and their rarity means that they serve very little real world application.

\paragraph{Training Configuration.}
All neural networks were trained using a 70-15-15 train/validation/test split. Predictor columns were scaled using a StandardScaler, which standardizes each feature to have zero mean and unit variance. Training used a batch size of 256 and validation and testing used a batch size of 512. All neural networks were trained using the Mean Absolute Error (MAE) loss function and the Adam optimizer with a learning rate of 1e-3 for 150 epochs. All tree-based models were trained using an 80-20 train/validation split, and when possible, were also trained with the Mean Absolute Error (MAE) loss function. For each of our hyperparameters, we tested at most 5 different values.

\paragraph{GNN Setup}
 We added BatchNorm \cite{Ioffe2015BatchNA} and residual connections \cite{He2015DeepRL} in between each layer to stabilize gradients and make training smoother. Furthermore, we used Jumping Knowledge \cite{Xu2018RepresentationLO} to improve model performance. All the GNNs were trained for 500 epochs.

\paragraph{Evaluation Metrics}
We use MAE and MSE as the evaluation metrics. MSE is the standard loss function for regression-related tasks, although it can easily be skewed by outlier bad predictions. MAE places less weight on outlier bad predictions, which is desirable as we believe that there will be cases where there are high levels of arsenic not easily explained by environment factors due to volatility of arsenic levels.

Our experimental results in Table~\ref{tab:comparison} show that GNN variants perform competitively with traditional tree-based models. In particular, GAT, GCN, and GraphSAGE achieve log-scale MAE values comparable to the widely adopted XGBoost baseline. Moreover, GAT, GraphSAGE, and GIN surpass XGBoost in real-scale MAE, with GraphSAGE improving performance by approximately $4.2\%$. Although their overall metrics are slightly lower in certain cases, GNN variants demonstrate strong robustness to outliers and superior stability in variance explanation. In contrast, both multilayer perceptrons and random forests exhibit noticeably weaker fits to the data, highlighting their limited capacity to model complex spatial and nonlinear relationships.

\section{Analysis and Discussion}

\begin{figure*}[t]
    \centering
    \begin{subfigure}[t]{\columnwidth}
        \centering
        \includegraphics[width=\linewidth]{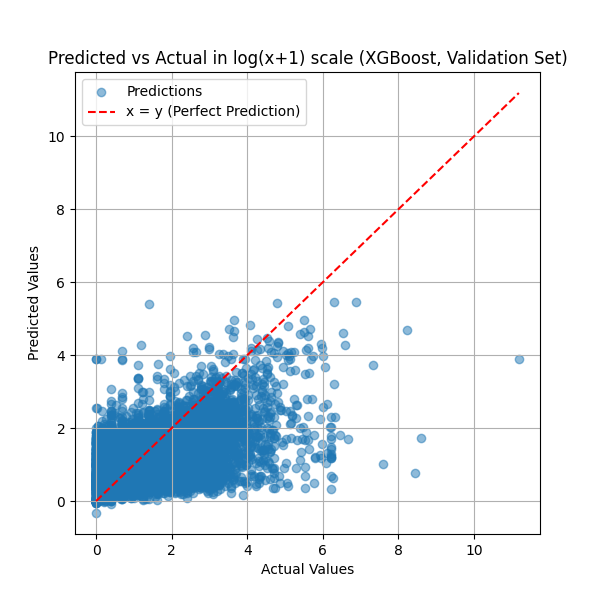}
        \caption{Predicted vs Actual in log(x+1) scale (XGBoost).}
        \label{fig:pred_vs_actual_xgboost}
    \end{subfigure}
    \vspace{0.5em}
    \begin{subfigure}[t]{\columnwidth}
        \centering
        \includegraphics[width=\linewidth]{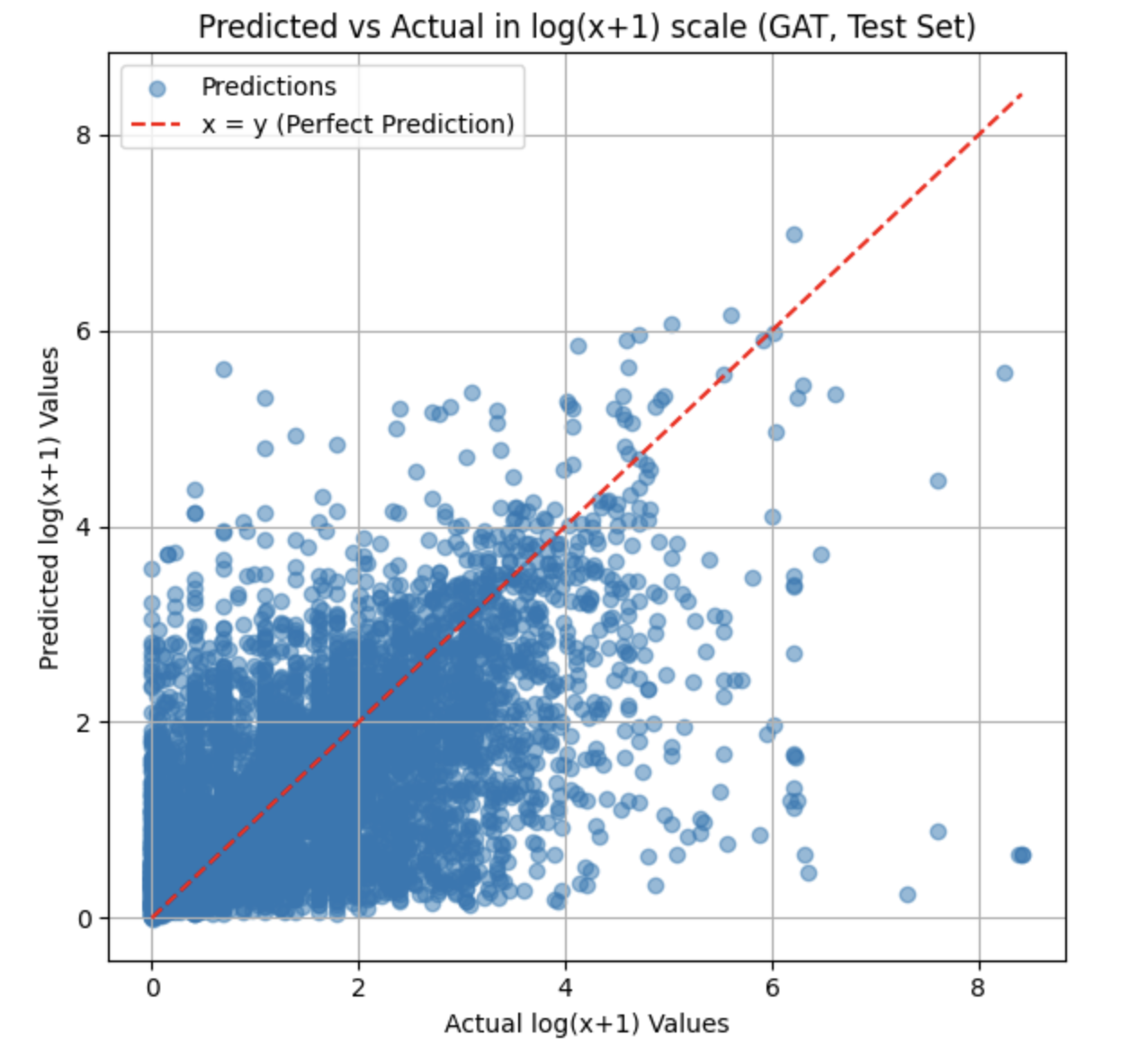}
        \caption{Predicted vs Actual in log(x+1) scale (GAT).}
        \label{fig:pred_vs_actual_gat}
    \end{subfigure}
    
    \caption{Predicted vs actual log levels of arsenic from XGBoost and GAT based models.}
    \label{fig:xgboost_combined}
\end{figure*}







GNNs are particularly powerful for this problem because they explicitly model spatial autocorrelation through graph-based learning. Since arsenic contamination exhibits strong spatial dependence, with latitude and longitude already being highly informative features, the GNN leverages these relationships by propagating information across connected locations. This enables it to capture spatial structure and inter-site dependencies more effectively than traditional models, improving upon previous studies \cite{Hu2022TheUO}.

We chose XGBoost to represent the gradient-boosted trees in error analysis. Figure~\ref{fig:xgboost_combined} shows that XGBoost tends to capture spatial and quantitative trends of arsenic pretty well, although its prediction errors do follow slight patterns. Because there are more points under the line y=x in Figure~\ref{fig:pred_vs_actual_xgboost}, we know that XGBoost tends to underestimate the amount of arsenic concentration. Figure~\ref{fig:spatial_error_xgboost} further supports this observation, as the residual distribution extends more strongly into the negative range and exhibits a higher density of blue (under predicted) points compared to red (over predicted) ones. This figure also indicates that XGBoost tends to over-predict or under-predict in states in the upper Midwest, such as Minnesota, Iowa, and Illinois.

We chose GAT to represent the graph neural networks in error analysis. Figure~\ref{fig:gat_combined} shows how GAT similarly tends to capture spatial and quantitative trends of arsenic pretty well, although patterns in its predictions are different from XGBoost. As shown in Figure~\ref{fig:pred_vs_actual_gat}, GAT's predictions are divided more evenly along the line y=x, which is most likely one of the big reasons it is able to slightly outperform the gradient-boosted trees. In this specific test set, there is a pretty big range of arsenic concentrations, and yet GAT is still able to estimate them pretty well. The spatial distribution error chart (Figure~\ref{fig:spatial_error_gat}) is very similar to XGBoost, with a large majority of errors being concentrated in the north Midwest. Besides occasionally over predicting values a bit more than XGBoost, the range of error between these two models are very similar.


\section{Conclusion and future work}
Avenues for future work can include integrating more types of environmental datasets and testing the performance of modern tabular transformers. Additionally, using classification to predict arsenic concentrations in water is also beneficial as it directly models the exceedance of regulatory thresholds. However, this approach is difficult due to the heavy class imbalance and the skewed data. GNNs are relatively unexplored in environmental science, and applying them to key environmental challenges represents an exciting research direction.

In this work, we assess a variety of machine learning models by investigating their ability to predict arsenic contamination in water across the contiguous United States. Our dataset also contains a variety of features aggregated from different public datasets, including substances in the soil and geologic data. Our findings suggest that GNN variants perform very well in environmental science tasks. By improving the robustness of arsenic prediction models while maintaining low loss, this study contributes to more accurate and reliable arsenic prediction.

\section{Acknowledgments}
The team is grateful for the researchers at Algoverse AI Research, particularly Kevin Han, for insightful discussions about methodology for this project.

\pretocmd{\bibliography}{\FloatBarrier}{}{}

\bibliography{aaai2026}

@article{Ayotte2017EstimatingTH,
    title={Estimating the High-Arsenic Domestic-Well Population in the Conterminous United States},
    author={Joseph D. Ayotte and Laura Medalie and Sharon L. Qi and Lorraine C. Backer and Bernard T. Nolan},
    journal={Environmental Science \& Technology},
    year={2017},
    url={https://api.semanticscholar.org/Corpus ID:6844338}
}

@article{Lombard2021MachineLM,
  title={Machine Learning Models of Arsenic in Private Wells Throughout the Conterminous United States As a Tool for Exposure Assessment in Human Health Studies.},
  author={Melissa A. Lombard and Molly Scannell Bryan and Daniel K. Jones and Catherine M. Bulka and Paul M. Bradley and Lorraine C. Backer and Michael J. Focazio and Debra T. Silverman and Patricia L Toccalino and Maria Argos and Matthew O. Gribble and Joseph D Ayotte},
  journal={Environmental science \& technology},
  year={2021},
  url={https://api.semanticscholar.org/CorpusID:232299884}
}

@article{Podgorski2020GlobalTO,
  title={Global threat of arsenic in groundwater},
  author={Joel E. Podgorski and Michael Berg},
  journal={Science},
  year={2020},
  volume={368},
  pages={845 - 850},
  url={https://api.semanticscholar.org/CorpusID:218836572}
}

@article{Subudhi2025IntegratingBL,
  title={Integrating Boosted learning with Differential Evolution (DE) Optimizer: A Prediction of Groundwater Quality Risk Assessment in Odisha},
  author={Sonalika Subudhi and Alok Kumar Pati and Sephali Bose and Subhasmita Sahoo and Avipsa Pattanaik and Biswa Mohan Acharya},
  journal={ArXiv},
  year={2025},
  volume={abs/2502.17929},
  url={https://api.semanticscholar.org/CorpusID:276580258}
}

@article{Hu2022TheUO,
  title={The Utility of Machine Learning Models for Predicting Chemical Contaminants in Drinking Water: Promise, Challenges, and Opportunities},
  author={Xindi C. Hu and Mona Q. Dai and Jennifer M. Sun and Elsie M. Sunderland},
  journal={Current Environmental Health Reports},
  year={2022},
  volume={10},
  pages={45 - 60},
  url={https://api.semanticscholar.org/CorpusID:254842780}
}

@article{Wu2021DistributionOG,
  title={Distribution of Groundwater Arsenic in Uruguay Using Hybrid Machine Learning and Expert System Approaches},
  author={Ruohan Wu and Elena Alvareda and David A. Polya and Gonzalo Blanco and Pablo Gamazo},
  journal={Water},
  year={2021},
  url={https://api.semanticscholar.org/CorpusID:233908670}
}

@article{Wu2023ArtificialIM,
  title={Artificial Intelligence Modelling to Support the Groundwater Chemistry-Dependent Selection of Groundwater Arsenic Remediation Approaches in Bangladesh},
  author={Ruohan Wu and Laura A. Richards and Ajmal Roshan and David A. Polya},
  journal={Water},
  year={2023},
  url={https://api.semanticscholar.org/CorpusID:264081317}
}

@article{Zhao2024PredictingAC,
  title={Predicting Arsenic Contamination in Groundwater: A Comparative Analysis of Machine Learning Models in Coastal Floodplains and Inland Basins},
  author={Zhenjie Zhao and Amit Kumar and Hongyan Wang},
  journal={Water},
  year={2024},
  url={https://api.semanticscholar.org/CorpusID:272168382}
}

@article{Chattopadhyay2023TheML,
  title={The machine learning and geostatistical approach for assessment of arsenic contamination levels using physicochemical properties of water.},
  author={Arghya Chattopadhyay and Anand Prakash Singh and Siddharth Kumar and Jayadeep Pati and Amitava Rakshit},
  journal={Water science and technology : a journal of the International Association on Water Pollution Research},
  year={2023},
  volume={88 3},
  pages={
          595-614
        },
  url={https://api.semanticscholar.org/CorpusID:260143927}
}

@article{Podgorski2020GroundwaterAD,
  title={Groundwater Arsenic Distribution in India by Machine Learning Geospatial Modeling},
  author={Joel E. Podgorski and Ruohan Wu and Biswajit Chakravorty and David A. Polya},
  journal={International Journal of Environmental Research and Public Health},
  year={2020},
  volume={17},
  url={https://api.semanticscholar.org/CorpusID:222166992}
}

@article{Tan2020MachineLM,
    title={Machine Learning Models of Groundwater Arsenic Spatial Distribution in Bangladesh: Influence of Holocene Sediment Depositional History},
    author={Zhen Tan and Qiang Yang and Yan Zheng},
    journal={Environmental Science and Technology},
    year={2020},
    url=
    {https://api.semanticscholar.org/Corpus ID:220474115}
}

@article{Zha2024GNN,
  author={Zha, Y. and Yang, Y.},
  title={Innovative graph neural network approach for predicting soil heavy metal pollution in the Pearl River Basin, China},
  journal={Scientific Reports},
  volume={14},
  pages={16505},
  year={2024},
  doi={10.1038/s41598-024-67175-7},
  publisher={Nature Publishing Group},
  url={https://doi.org/10.1038/s41598-024-67175-7}
}

@article{Truong2023GraphNN,
  title={Graph Neural Networks for Pressure Estimation in Water Distribution Systems},
  author={Huy Truong and Andr{\'e}s Tello and Alexander Lazovik and Victoria Degeler},
  journal={Water Resources Research},
  year={2023},
  volume={60},
  url={https://api.semanticscholar.org/CorpusID:265281555}
}

@article{Sun2022AGN,
  title={A graph neural network (GNN) approach to basin-scale river network learning: the role of physics-based connectivity and data fusion},
  author={Alexander Y. Sun and Peishi Jiang and Zong‐Liang Yang and Yangxinyu Xie and Xingyuan Chen},
  journal={Hydrology and Earth System Sciences},
  year={2022},
  url={https://api.semanticscholar.org/CorpusID:252943700}
}

@article{Kipf2016SemiSupervisedCW,
  title={Semi-Supervised Classification with Graph Convolutional Networks},
  author={Thomas Kipf and Max Welling},
  journal={ArXiv},
  year={2016},
  volume={abs/1609.02907},
  url={https://api.semanticscholar.org/CorpusID:3144218}
}

@article{Velickovic2017GraphAN,
  title={Graph Attention Networks},
  author={Petar Velickovic and Guillem Cucurull and Arantxa Casanova and Adriana Romero and Pietro Lio’ and Yoshua Bengio},
  journal={ArXiv},
  year={2017},
  volume={abs/1710.10903},
  url={https://api.semanticscholar.org/CorpusID:3292002}
}

@article{Hamilton2017InductiveRL,
  title={Inductive Representation Learning on Large Graphs},
  author={William L. Hamilton and Zhitao Ying and Jure Leskovec},
  journal={ArXiv},
  year={2017},
  volume={abs/1706.02216},
  url={https://api.semanticscholar.org/CorpusID:4755450}
}

@article{Xu2018HowPA,
  title={How Powerful are Graph Neural Networks?},
  author={Keyulu Xu and Weihua Hu and Jure Leskovec and Stefanie Jegelka},
  journal={ArXiv},
  year={2018},
  volume={abs/1810.00826},
  url={https://api.semanticscholar.org/CorpusID:52895589}
}

@article{Chen2016XGBoostAS,
  title={XGBoost: A Scalable Tree Boosting System},
  author={Tianqi Chen and Carlos Guestrin},
  journal={Proceedings of the 22nd ACM SIGKDD International Conference on Knowledge Discovery and Data Mining},
  year={2016},
  url={https://api.semanticscholar.org/CorpusID:4650265}
}

@inproceedings{Ke2017LightGBMAH,
  title={LightGBM: A Highly Efficient Gradient Boosting Decision Tree},
  author={Guolin Ke and Qi Meng and Thomas Finley and Taifeng Wang and Wei Chen and Weidong Ma and Qiwei Ye and Tie-Yan Liu},
  booktitle={Neural Information Processing Systems},
  year={2017},
  url={https://api.semanticscholar.org/CorpusID:3815895}
}

@inproceedings{Ostroumova2017CatBoostUB,
  title={CatBoost: unbiased boosting with categorical features},
  author={Liudmila Ostroumova and Gleb Gusev and Aleksandr Vorobev and Anna Veronika Dorogush and Andrey Gulin},
  booktitle={Neural Information Processing Systems},
  year={2017},
  url={https://api.semanticscholar.org/CorpusID:5044218}
}

@article{Ioffe2015BatchNA,
  title={Batch Normalization: Accelerating Deep Network Training by Reducing Internal Covariate Shift},
  author={Sergey Ioffe and Christian Szegedy},
  journal={ArXiv},
  year={2015},
  volume={abs/1502.03167},
  url={https://api.semanticscholar.org/CorpusID:5808102}
}

@article{He2015DeepRL,
  title={Deep Residual Learning for Image Recognition},
  author={Kaiming He and X. Zhang and Shaoqing Ren and Jian Sun},
  journal={2016 IEEE Conference on Computer Vision and Pattern Recognition (CVPR)},
  year={2015},
  pages={770-778},
  url={https://api.semanticscholar.org/CorpusID:206594692}
}

@article{Xu2018RepresentationLO,
  title={Representation Learning on Graphs with Jumping Knowledge Networks},
  author={Keyulu Xu and Chengtao Li and Yonglong Tian and Tomohiro Sonobe and Ken-ichi Kawarabayashi and Stefanie Jegelka},
  journal={ArXiv},
  year={2018},
  volume={abs/1806.03536},
  url={https://api.semanticscholar.org/CorpusID:47018956}
}

@misc{environmental_protection_agency_water_2013,
	title = {Water Quality Portal},
	url = {https://www.waterqualitydata.us/},
	doi = {10.5066/P9QRKUVJ},
	publisher = {U.S. Geological Survey},
	author = {{Environmental Protection Agency} and {United States Geological Survey}},
    howpublished = "\url{https://www.waterqualitydata.us/}",
    note="Accessed: 2025-10-23",
	urldate = {2025-10-23},
	year = 2013,
}

@misc{soil_survey_staff_gridded_2025,
	title = {Gridded National Soil Survey Geographic ({gNATSGO}) Database for the Conterminous United States},
	url = {https://nrcs.app.box.com/v/soils},
	titleaddon = {Department of Agriculture, Natural Resources Conservation Service},
	author = {Soil Survey Staff},
	urldate = {2025-10-23},
    year = 2025,
	date = {2025-10-23},
	langid = {english},
}

@techreport{mason_mineral_1996,
	title = {Mineral Resources Data System ({MRDS})},
	url = {https://pubs.usgs.gov/publication/ds20},
	number = {20},
	institution = {The Survey,},
	author = {Mason, G. T. and Arndt, R. E.},
    year = 1996,
	urldate = {2025-10-23},
	langid = {english},
	doi = {10.3133/ds20},
	note = {{ISBN}: 9780607855081
{ISSN}: 2327-638X
Publication Title: Data Series},
}

\end{document}